\documentclass{article}

\usepackage{arxiv}

\usepackage[utf8]{inputenc} 
\usepackage[T1]{fontenc}    
\usepackage{hyperref}       
\usepackage{url}            
\usepackage{booktabs}       
\usepackage{amsfonts}       
\usepackage{nicefrac}       
\usepackage{microtype}      
\usepackage{lipsum}
\usepackage{graphicx}
\usepackage{caption}
\graphicspath{ {./images/} }

\title{The AI Arena: A Framework for Distributed Multi-Agent Reinforcement Learning }
\author{Edward W. Staley\footnotemark[1] ,
Corban G.Rivera\footnotemark[1] ,
Ashley J. Llorens\footnote{Equal contribution}\\
Intelligent Systems Center\\
Johns Hopkins Applied Physics Laboratory\\

}

\begin{document}
\maketitle
\begin{abstract}
Advances in reinforcement learning (RL) have resulted in recent breakthroughs in the application of artificial intelligence (AI) across many different domains. An emerging landscape of development environments is making powerful RL techniques more accessible for a growing community of researchers. However, most existing frameworks do not directly address the problem of learning in complex operating environments, such as dense urban settings or defense-related scenarios, that incorporate distributed, heterogeneous teams of agents. To help enable AI research for this important class of applications, we introduce the AI Arena: a scalable framework with flexible abstractions for distributed multi-agent reinforcement learning. The AI Arena extends the OpenAI Gym interface to allow greater flexibility in learning control policies across multiple agents with heterogeneous learning strategies and localized views of the environment. To illustrate the utility of our framework, we present experimental results that demonstrate performance gains due to a distributed multi-agent learning approach over commonly-used RL techniques in several different learning environments. 

\end{abstract}

\section{INTRODUCTION}

Reinforcement learning (RL) offers a powerful approach to generating complex behaviors for intelligent systems that could not be explicitly derived or programmed. In the RL setting, the problem of learning an effective control policy is posed as a sequential decision-making problem for an agent interacting with a learning environment \cite{sutton:RL}. Given that learning the environment dynamics is an essential aspect of the RL problem, the ultimate effectiveness of a learned policy is dependent on the extent to which the learning environment reflects the essential aspects of the intended operating environment for the target system. Hence, many RL breakthroughs to date have focused on gaming and other applications with structured and predictable environments \cite{alphastar,alphazero,poker}. 

Translating progress in RL to increasingly complex applications of artificial intelligence (AI) will require the design of representative learning environments with corresponding complexity. Ensuring that future progress is reproducible and accessible for a broad community of researchers will require tools and frameworks that enable RL solutions to gracefully scale to address the problem of learning effectively in these increasingly complex settings. In general, RL frameworks must balance multiple tradeoffs, including ease of prototyping versus training at scale, high-level abstractions versus fine-grained control, and richness of features versus ease of use.

In this work, we introduce the AI Arena: a scalable framework with flexible abstractions for distributed multi-agent reinforcement learning. Our aim is to help enable RL research for the class of applications that involve multiple teams of agents where each team may have unique learning strategies and where agents within a given team may have localized views of the environment. Distributed multi-agent applications may be thought of as analogous to "system of systems" applications from a systems engineering perspective where collections (teams) of goal-oriented systems (agents) collaborate to achieve shared objectives. These attributes may arise, for example, in smart city applications where automated traffic control systems interact with fleets of automated vehicles or in defense applications where heterogeneous autonomous systems interact across time and space to achieve high-level mission objectives. Applications such as these often include cooperation or competition among heterogeneous teams of agents as defining features.

The AI Arena extends the commonly-used OpenAI Gym \cite{gym} interface to provide a flexible and scalable RL framework for the class of distributed multi-agent applications described above. Specifically, the AI Arena offers:

\begin{itemize}
    \item \textit{Multi-policy learning}: Each team of agents in the environment may employ a unique and potentially decentralized learning strategy. 
    \item \textit{Distributed learning}: Each agent in a team can be instantiated as a worker deployed to an arbitrary computing resource contributing learning experiences to a common control policy.
    \item \textit{Flexible curriculum design}: Successive learning episodes can be easily structured as sequences of self-play, tournament play or population-based policy selection.
    \item \textit{Broad compatibility}: Support for a wide range of different simulation environments and commonly-used base learning strategies.
\end{itemize}

The primary contribution of this work is the introduction of the AI Arena framework along with a set of motivating experimental results. To illustrate the efficacy of the our framework, we apply known algorithms to multi-agent problems and demonstrate how our framework can easily be used to train against these problems at scale. We present experimental results of single-agent algorithmic approaches to a multi-team competitive environment called TanksWorld \cite{tanksworld}, for which each environment instances houses 10 agents. We further demonstrate how multi-agent algorithms are supported by our framework by presenting results from the Cooperative Navigation environment \cite{maddpg}.


\section{RELATED WORK}
The OpenAI Gym \cite{gym} interface have been widely adopted across RL research frameworks, especially those primarily focused on single-agent environments \cite{baselines,spinningup,stablebaselines}. The AI Arena builds upon the standard Gym interface to provide robust support for multi-agent environments with decentralized control. Google Dopamine \cite{dopamine} is a highly configurable RL framework via for supports dynamic function declaration. KerasRL \cite{kerasrl} includes a good diversity of algorithms including DDPG, DQN, and SARSA focused on single agent reinforcement learning. Deepmind TRFL includes reinforcement learning primitives that can be direly integrated with tensorflow, while other neural network libraries are not well supported. Tensorforce \cite{tensorforce} also integrates strongly with tensorflow and includes a good abstraction for defining a round of training. The strong integration with tensorflow is also a limitation when considering reinforcement learning algorithms that are not based on deep networks. Facebook's Horizon \cite{horizon} framework has extensive support for distributed and scaled training through Spark. Horizon is focused on the scaled use case and is relatively more verbose when prototyping. The framework has a large amount of external dependencies and the focus is primarily on off-policy reinforcement learning. 

Coach \cite{coach} has a large number of reinforcement learning algorithms built-in. It has native support for scaling through Kubernetes. TF-agents \cite{tfagents} aims for production level training. The framework focuses on ease of prototyping and it is tightly coupled to tensorflow. SLM-Lab \cite{slmlab} focuses on designing reproducible experiments for the comparison of approaches. The framework uses Ray for scalability and focuses on single agent training via pytorch-based neural networks. DeeR is another deep reinforcement learning framework with solid abstractions and algorithm support. 
Garage is derived from rllib and implements a large number of reinforcement learning, meta-RL, and imitation learning algorithms. Surreal \cite{surreal} is a reinforcement learning framework with built-in support for robotics simulation via Mujoco. RLgraph \cite{rlgraph} focuses on scalability also using Ray \cite{ray}. It includes abstractions that avoid coupling the RL algorithm with the framework. ChainerRL \cite{chainerrl} also includes a broad set of reinforcement learning algorithms. The framework is specifically tailored for the use of the Chainer library for deep networks. Some libraries \cite{simplerl,mushroomrl,rlpyt} focus on reproducibility of experiments and ease of prototyping. These frameworks contain a tremendous amount of capability, but only specialize in single agent learning tasks. They lack abstractions suitable for multi-agent multi-policy learning, which is the focus of this work.

Related to our framework, MAgent \cite{magent} has support for large numbers of multi-agent training including multiple policy types. The abstraction for registering agents and agent types is useful. A weakness is the built-in focus on gridworld like environments. It might be challenging to integrate other environments into the framework. RLlib \cite{rllib} is part of the Ray project. It includes relatively easy access to distributed training through Ray. The library includes solid abstractions for multi-agent multi-policy training, and the configuration in python code is relatively easy to use. RLlib includes Tune which provides convenient access to built-in hyperparameter tuning. However, the framework has relatively high complexity, and integrating new algorithms can be challenging as native PyTorch or Tensorflow models must be wrapped in the RLlib Model abstraction. ML-agents \cite{mlagents} includes good support for multi-agent multi-policy training. It promotes flexible coupling between agents and policies. However, the framework is tightly coupled to the Unity simulation engine, and it currently lacks multi-node distributed training.

\section{AI ARENA INTERFACE}
\subsection{INTERFACE OVERVIEW}

\begin{figure}[ht!]
\centering
  \includegraphics[width=1.0\linewidth]{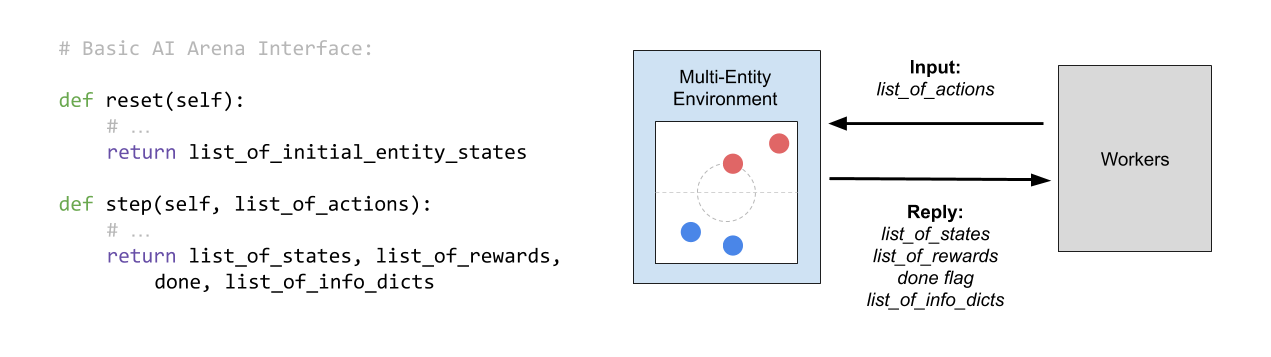}
  \captionsetup{width=0.95\linewidth}
  \caption{\small \emph{The AI Arena Interface.} The AI Arena Interface is an adjustment to the OpenAI Gym interface such that many data objects are replaced with lists. This is easily added to an existing environment's reset() and step() methods such that a list of actions is accepted and lists of response data are returned.}
  \label{fig:figted1}
\end{figure} 

The AI Arena Interface is an extension of the gym interface \cite{gym} that allows for environments to house multiple agents, without limitation on how those agents are attached to learning algorithms. This revision to the gym interface causes a distinct change in thinking about how environments are constructed: the environment is no longer completely representative of the problem a given agent is trying to solve. Rather, the environment is a shared space among one or more entities, each potentially having their own objectives. We use the term “entity” to avoid confusion between a policy and the portion of the environment under its control.

The above interface changes are manifested by simply expecting lists of values where the gym interface expects singular values. The only exception is the “done” flag, which is kept as a single boolean to indicate if the shared episode has ended. This data flow is described in Figure \ref{fig:figted1}.

\subsection{MULTIAGENT ENVIRONMENTS} 
One primary goal of the AI Arena interface is to encourage environments in which a variety of agents may coexist and learn together. This should encompass everything from collaboration among identical entities to competition among several dissimilar groups. To that end, other properties of the environment are also converted to lists, such as action spaces or observation spaces. This allows for a variety of agent types to coexist in a single environment. For example, one learning entity may be making discrete decisions about image state data, while another entity in the same environment may expect continuous actions based on a vectorized state space.

An implication of this multi-entity setup is that all entities, as well as their actions, observations, and rewards, are occurring in lock-step at the same rate. Each step, the environment expects decisions corresponding to all entities, and will return information to all of them about the consequences of those decisions. While this may seem limiting at first, it is better to think of this as supporting the most extreme case of multi-agent interaction: all entities can be involved in a single frame of the environment. It is fairly straightforward to embed special cases within this framework: an entity which has exited an episode early can send and receive null values, or an entity with a lower interaction frequency can easily on every $N$th observation and repeat actions until that observation occurs. The global “done” signal is especially useful for simulations and games in which there is a common or mutually exclusive objective, as is often the case. 

\subsection{MULTIPLE LEARNING POLICIES}
\begin{figure}[ht!]
\centering
  \includegraphics[width=\linewidth]{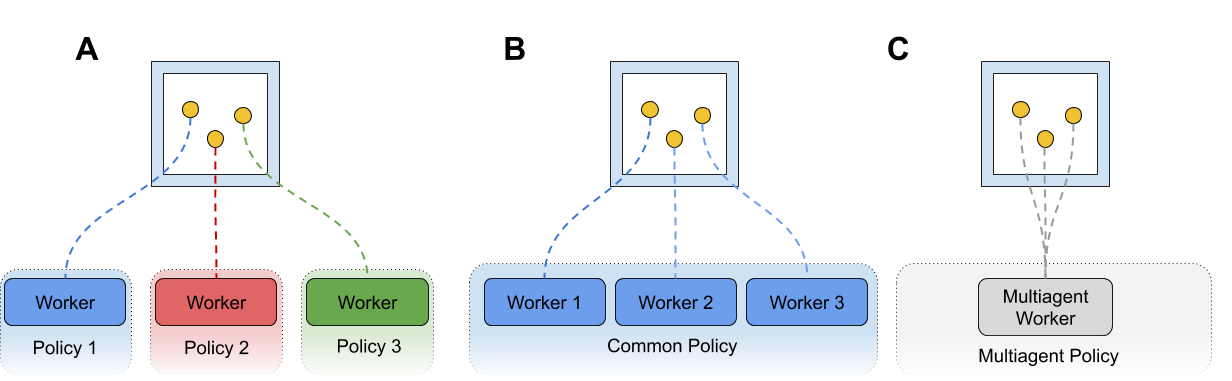}
  \captionsetup{width=0.95\linewidth}
  \caption{\small \emph{Possible worker configurations.} Policy workers may be attached in environment entities in any desired combination. \textbf{A} Three entities are assigned to three independent policies that are learning separately and may have workers attached to other environments. \textbf{B} Entities are each attached to workers of the same policy, such that some of the agents contributing to the policy are coexisting in the same environment. \textbf{C} Entities are all attached to the same policy worker, which takes all of their data into account in a multiagent manner, possibly one of many workers for a distributed multiagent algorithm.}
  \label{fig:figted2}
\end{figure}

A further goal of the AI Arena interface is to enable complex distributed training architectures in which many policies may be training simultaneously in shared environments. The policies may be several instances of the same algorithm or be entirely separate approaches to learning. The inclusion of many entities in a single environment breaks from a typical training paradigm of one policy-worker thread corresponding to one agent in one environment. Rather, it is up to the user of this interface to distribute the many entities in an environment (or across many environments in the distributed case) to as many agents as desired. For example, an environment with $N$ agents may function as $N$ workers to a single distributed algorithm, or on the other extreme, single workers to $N$ distinct policies. They may also be grouped such that $M$ agents are in fact controlled by a single instance of a multi-agent policy (Figure \ref{fig:figted2}). In other words, the agency of a given entity is at the discretion of the user. While this is a potentially powerful paradigm, it can be complex to implement. In the next section, we will describe the AI Arena software framework, which provides a straightforward implementation.

\section{AI ARENA SOFTWARE FRAMEWORK} 
The AI Arena software framework is an implementation of the previously described AI Arena Interface with several goals in mind: 1) To allow usage of any environment or RL algorithm that the user may desire, with minimal effort, 2) To provide high-level functionality such that a user can quickly describe a very sophisticated training or testing scheme in python, and 3) to seamlessly orchestrate the desired training at scale.

\subsection{PROCESS-BASED ARCHITECTURE}
\begin{figure}[ht!]
\centering
  \includegraphics[width=0.7\linewidth]{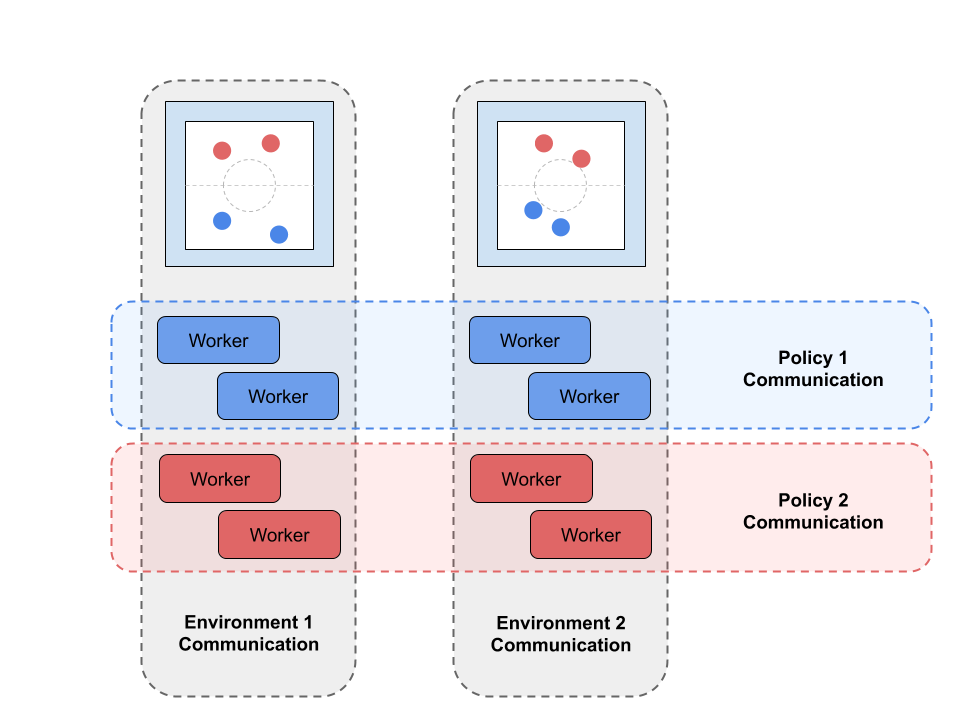}
  \captionsetup{width=0.95\linewidth}
  \caption{\small \emph{Communication groupings among AI Arena processes.} MPI communication occurs between each environment and any policy workers which are commanding its entities. Separately, workers with a common policy communicate to share network weights, experience, or gradient information.}
  \label{fig:figted3}
\end{figure}

The AI Arena software framework is designed to run in a highly distributed and compartmentalized way, making heavy use of processes via the Message-Passing Interface (MPI) standard \cite{openmpi,mpi4py}. Each environment instance is maintained in its own process, as well as each policy worker. The MPI paradigm allows many processes to spin up at once (across one or more machines), and allows various subsets of processes to communicate. This is used by the AI Arena to designate messaging groups between an environment and any policy workers interfacing with that environment, as well as between collections of worker processes which all contribute to a single policy. These groupings are described in Figure \ref{fig:figted3}, and details on environment and algorithm integration in the following subsections.

The AI Arena uses this grouping structure to orchestrate training runs with multiple environments, each having multiple entities, and contributing to multiple policies. This constitutes one training “rounds”. While many reinforcement learning approaches train for an indeterminate amount of time, the AI Arena also handles the bringing down of distributed processes at specific times. This allows many rounds to be conducted seamlessly in sequence, enabling training schemes that progress through tournament structures, evolution of populations of agents, or schedules of environment parameters.

\subsection{ENVIRONMENT INTEGRATION}
As described previously, each environment is contained in a dedicated CPU process, and MPI is used to communicate between an environment and external agents. However, these details are fully obscured from a user looking to attach an environment. Each environment is wrapped in a python class which manages the communication to and from the environment, as well as MPI communication to and from policy workers. Actions from every relevant process are aggregated and sent to the worker as a single, local list communication, and replies are received as a list and then disseminated back to the respective parties (Figure \ref{fig:figted3} left). From the environment design perspective, the environment only needs to implement the previously described AI Arena Interface.

It is not uncommon for a python environment instance to itself manage an external environment, and the AI Arena places no restrictions on this design. For example, once an environment process is created by the AI Arena, an environment instance will be created. At this point, the python environment may bring up, for example, a Unity simulation which has its own lines of communication. This communication is completely separated from the AI Arena framework, just as the MPI calls are completely separated from the python environment instance.

\begin{figure}[ht!]
\centering
  \includegraphics[width=\linewidth]{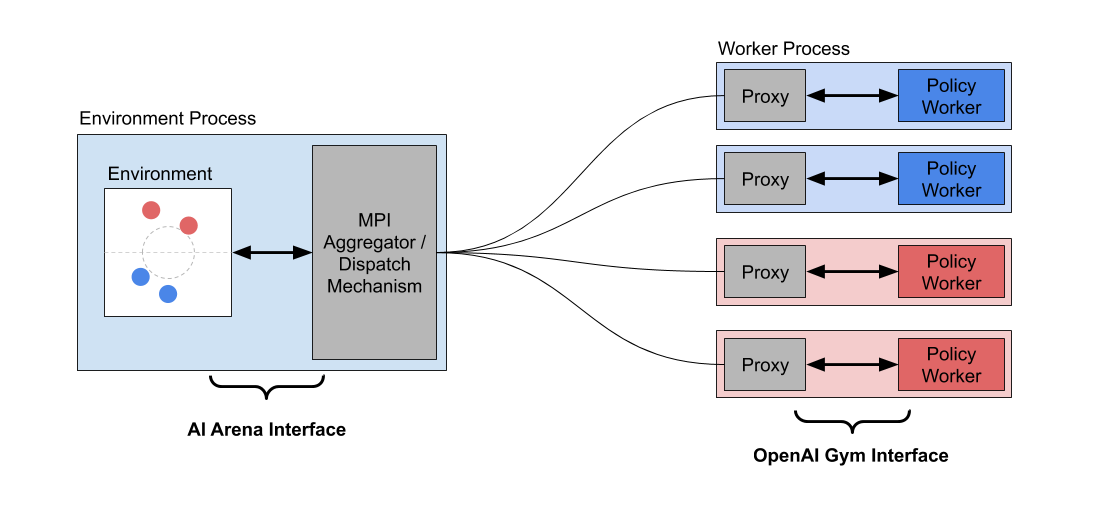}
  \captionsetup{width=0.95\linewidth}
  \caption{\small \emph{Communication between an environment and single-agent policy workers.} \textbf{Left:} An environment instance is contained in a process which manages MPI communication. The core environment only needs to manage lists of actions and lists of replies. \textbf{Right:} Single-agent policy workers communicate locally with an environment proxy that handles MPI communication to a specific entity, expressed as a single-agent gym interface. Existing single-agent algorithms can be easily integrated using this paradigm.}
  \label{fig:figted4}
\end{figure} 

\subsection{ALGORITHM INTEGRATION}
Just as the environment instances are contained in unique CPU processes, so are workers contributing to the training of a distributed reinforcement learning policy. MPI is again used here to maintain communication among a set of workers, to share gradients, data, or network parameters. Distributed algorithms looking to train with the AI Arena framework are encouraged to be set up in this manner, using MPI for updates as in \cite{spinningup}. However, other forms of communication could be used among processes at the user’s discretion.

While the AI Arena Interface encourages highly multi-entity environments, many DRL algorithms are constructed to be single agent, and to communicate with the standard gym interface. To bridge this gap, the AI Arena software framework provides to each worker a proxy environment which implements the typical single-agent OpenAI gym interface, and communicates with the true environment on the backend. In this manner, a single-agent policy can be run without any modification. From its perspective, it is operating a single agent in a single environment, and may be unaware that other entities or other policies exist alongside it (Figure \ref{fig:figted3} right).

While this provides simple integration for many problems, it is not always sufficient to apply a single-agent algorithm to a multi-agent environment, and multi-agent algorithms have been developed for that use case. The AI Arena fully supports the assignment of multiple entities to a single policy worker rather than one, and in this case the worker will receive lists of data about several entities and be expected to reply with lists of actions. This enables, for example, a single worker process to manage all of the entities within an environment (or some subset). Scaling up to many environments, this would simply result in distributed training of a multi-agent algorithm. Furthermore, it is perfectly compatible to mix multi-entity and single-entity assignments, such that some entities may be attached to single-agent algorithms while others are pooled into multiagent algorithms.

\begin{figure}[ht!]
\centering
  \includegraphics[width=\linewidth]{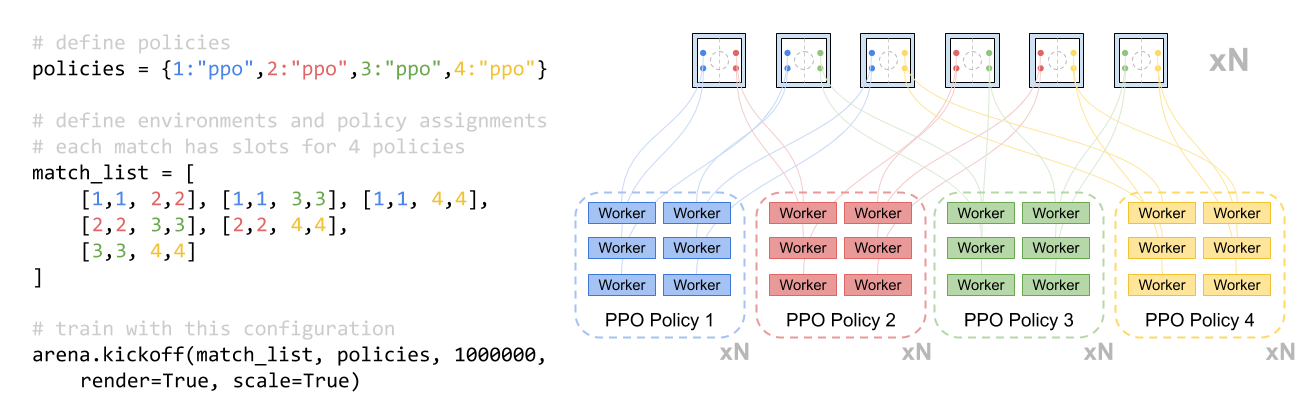}
  \captionsetup{width=0.95\linewidth}
  \caption{\small \emph{Python code and process diagram for a four-policy round robin training scheme.} A hypothetical training scheme in which four policies train simultaneously in a round-robin style with a 2v2 environment. \textbf{Left:} Given an environment that conforms to the AI Arena Interface, the core content of the script to run this training scheme is only three lines. Four policies are defined, arranged into six environments, and the Arena is asked to run the provided configuration. \textbf{Right:} The resulting processes and their basic organization. With scaling turned on, this brings up 30$N$ processes to run the configuration $N$ times over simultaneously.}
  \label{fig:figted5}
\end{figure}

\subsection{TRAINING DESCRIPTIONS AND SCALABILITY}
Finally, the AI Arena software framework is designed to orchestrate much of the above implementation without any input from the user of the framework. Training schemes that previously would be quite cumbersome to maintain can be described in only a few lines of code (Figure \ref{fig:figted5} left). Furthermore, the AI Arena software framework includes many built-in utilities and conveniences, including example environments and common algorithms. Notably, a complex training scheme can be automatically duplicated across available compute. For example, the round-robin setup described in Figure \ref{fig:figted5} can be automatically repeated $N$ times over such that $6N$ environments exist, connected to $24N$ workers representing 4 policies (Figure \ref{fig:figted5} right). Policy hooks allow for automatic logging, saving, and restoration to ease the maintenance of such complex setups. Additional utilities allow for policies to be duplicated so that they may be developed over several rounds in population-based approaches.


\section{RESULTS AND DISCUSSION}

In this section we highlight results that illustrate some of the key features of the AI Arena. To that end, we demonstrate the use of multi-agent, multi-policy, distributed curriculum training in the AI Safety Tanksworld domain \cite{tanksworld}, and demonstrate options for multi-agent algorithms in a cooperative navigation domain \cite{maddpg}. Additional examples that demonstrate advanced training schemes made possible by the AI Arena are included in the appendix.

\subsection{AI SAFETY TANKSWORLD DOMAIN}

AI Safety TanksWorld is an environment for AI safety research with three essential aspects: competing performance objectives, human-machine teaming, and multi-agent competition. The AI Safety TanksWorld aims to accelerate the advancement of safe multi-agent decision-making algorithms by providing a software framework to support competitions with both system performance and safety objectives \cite{tanksworld}.

\begin{figure}[ht!]
\centering
  \includegraphics[width=\linewidth]{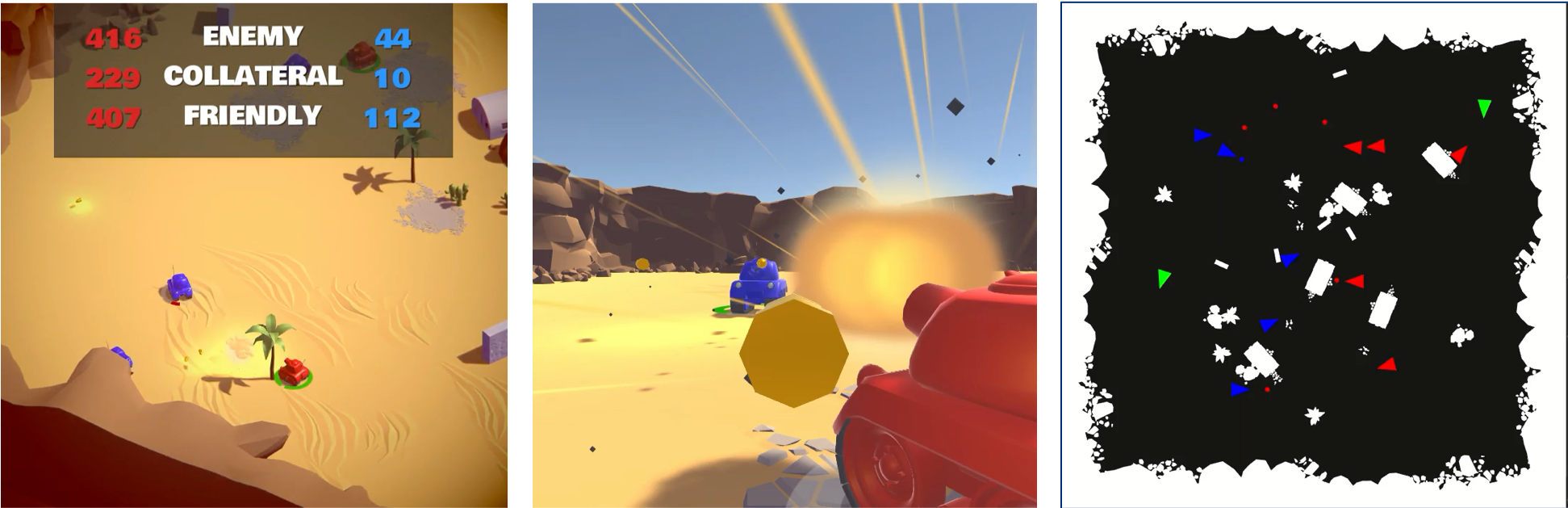}
  \caption{TanksWorld Multi-Agent Environment for AI Safety Research. These images illustrate different views of the Tanksworld environment (Left) is a birds-eye rendering of the environment, (Center) is an agent's-view rendering, and (Right) is the state representation actually provided as observations to the RL algorithm }
  \label{fig:tanksworld}
\end{figure} 

More specifically, the AI Safety TanksWorld \cite{tanksworld} is a competitive 5 vs. 5 environment that challenges teams of agents to simultaneously win against the opposing team, cooperate with diverse teammates including humans, and cope with uncertainty in the environment.

\begin{figure}[ht!]
\centering
  \includegraphics[width=\linewidth]{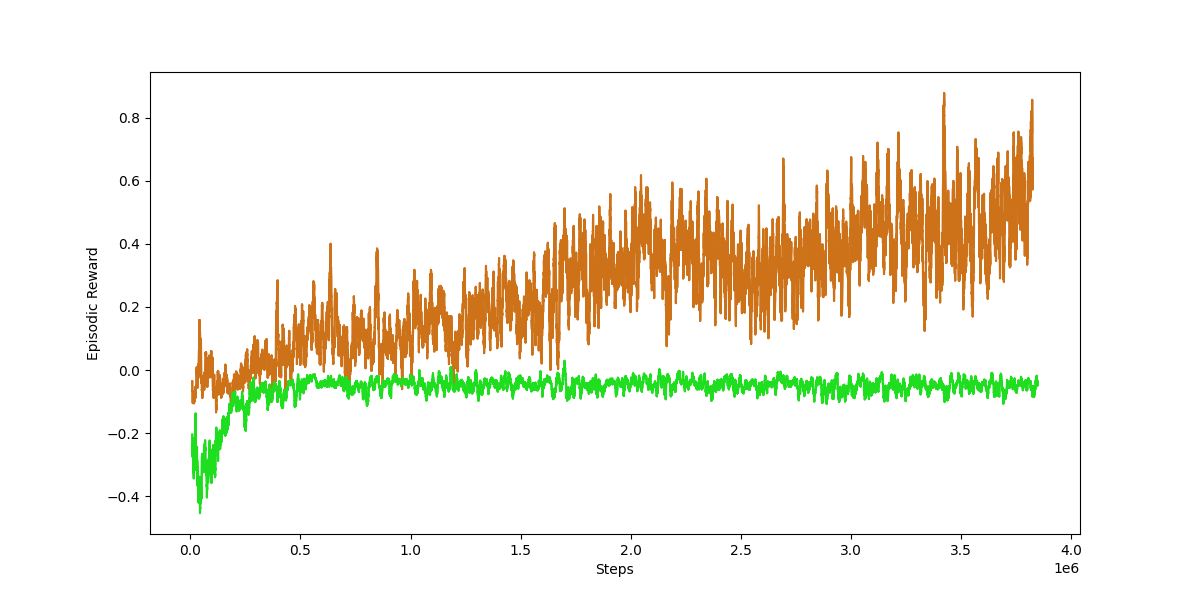}
  \captionsetup{width=0.95\linewidth}
  \caption{\small Round-based curriculum training organized with the AI Arena (brown). Successive rounds of training increased the difficulty by slowing introducing safety penalties over three rounds of training with penalty weights (0,.05,.1,.15,.20,.25,.3). The baseline (green) starts with the penalty threshold of $.3$. The result illustrates the value of successive rounds of curriculum training for teams of tanks in the AI Safety Challenge domain }
  \label{fig:tanksworld_result}
\end{figure}

\begin{figure}[ht!]
\centering
  \includegraphics[width=\linewidth]{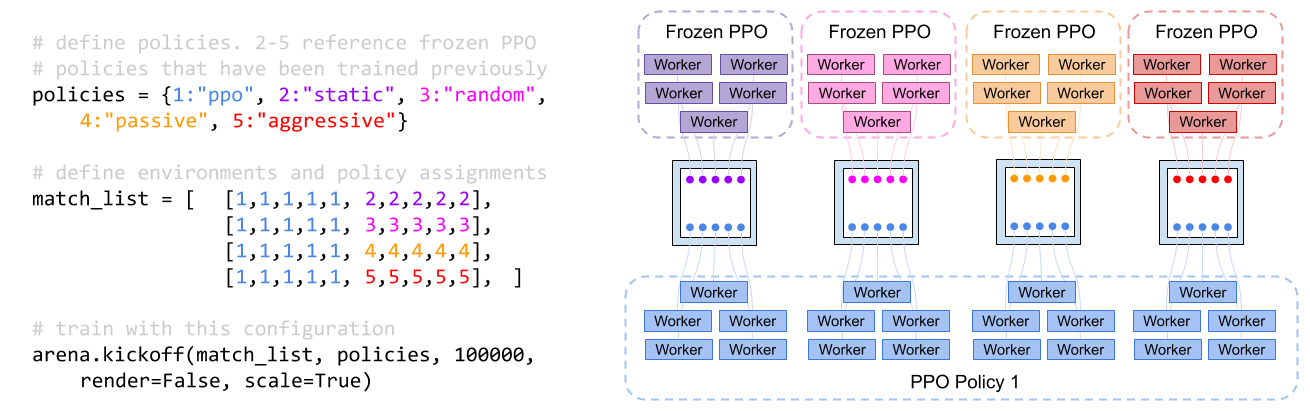}
  \captionsetup{width=0.95\linewidth}
  \caption{\small \emph{Python code and process diagram for a TanksWorld training scheme against four enemy strategies.} \textbf{Left:} Core code snippet needed to setup and run this training scheme. "ppo" refers to a new ppo policy that is being trained, while the other policies refer to custom frozen policies. \textbf{Right:} The resulting processes and their organization. Four environments are created, each housing 10 entities. All blue tanks are contributing to a single policy.}
  \label{fig:tanks_code}
\end{figure}

To illustrate the expressiveness of the AI Arena for complex reinforcement learning experimental design, we organized an experiment that compares  reinforcement learning training against multiple opponents simultaneously with and without curriculum training. As shown in Figure \ref{fig:tanks_code}, we train against 4 different opponent policies (i.e. static, random, aggressive, and passive policies).    The policy weights for the aggressive and passive policies were pretrained via PPO and frozen. Curriculum training was used to slowly introducing penalties for safety violations. The curriculum was composed of increasing penalties for safety violations from 0 to .3 in increments of .05 distributed evenly over 4 million steps  We compared the curriculum training approach to a baseline approach without a curriculum that sets the penalty for safety violations at .3.  The results of the comparison are shown in Figure \ref{fig:tanksworld_result}.  The non-curriculum baseline reaches at plateau at just below 0, while the curriculum-based approach achieves a higher overall combined score for both safety and performance.  The accompanying code shown in Figure \ref{fig:tanks_code} illustrates the expressiveness of the AI Arena abstractions for multi-agent training.

\begin{figure}[ht!]
\centering
  \includegraphics[width=.5\linewidth]{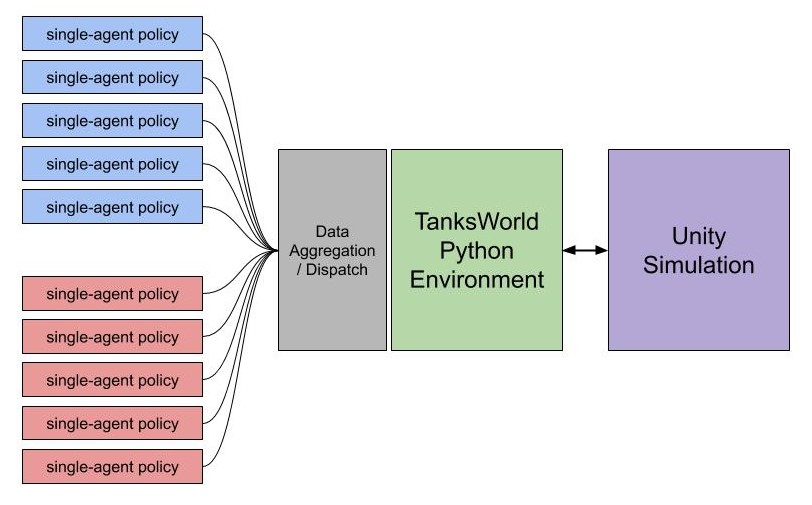}
  \captionsetup{width=0.95\linewidth}
  \caption{\small The AI Safety TanksWorld is composed of 10 controllable entities divided into two teams. The AI Arena is used to facilitate training of separate policies for each teams, and in this example all 5 team members use and train the same policy.}
  \label{fig:multiagent_tanksworld}
\end{figure} 

\subsection{Multi-agent Algorithms}

When working with environments containing multiple agents, it is sometimes desirable to coordinate their behavior beyond treating them as copies of single-agent policies. A recommended approach in this case is to use a common critic architecture, as introduced in multi-agent deep distributed policy gradient \cite{maddpg}, which uses an understanding of collective rewards during training to create individual policies that work together. To support such an approach with the AI Arena, multiple entities can be assigned to a single policy worker, such that a single worker coordinates the behavior of multiple entities.

\begin{figure}[ht!]
\centering
  \includegraphics[width=\linewidth]{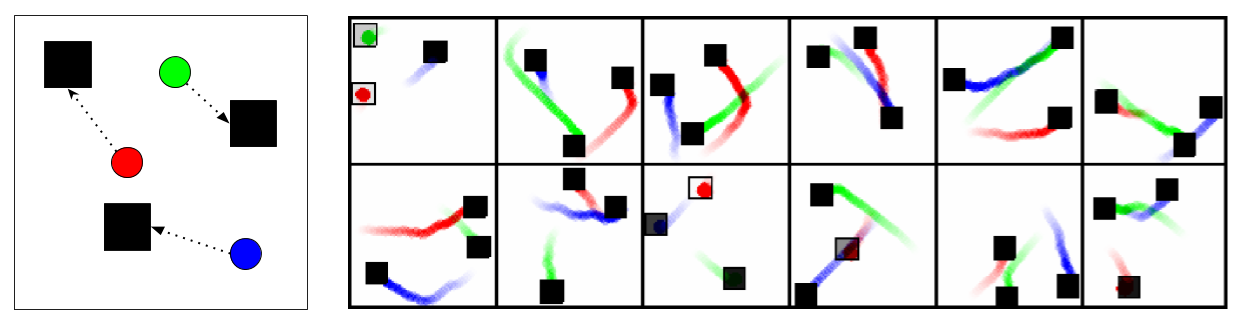}
  \captionsetup{width=0.95\linewidth}
  \caption{\small \emph{Cooperative Navigation Environment and Behavior.} The Cooperative Navigation environment has three targets (black squares) and three agents (circles). The agents must coordinate to cover all targets. \textbf{Left:} Illustration of environment and potential solution. \textbf{Right:} Traces of the testing behavior from the learned MASAC policies. The actions were reduced in magnitude to create slow paths to the targets. Some targets appear as outlines, showing that the agent happened to start on or near that target. The traces are interesting in that they show clear coordination but occasionally sub-optimal pairings of entities and targets. If the actions were not reduced, such that the entities reached the targets in only a handful of steps, these sub-optimalities would have little consequence on score.}
  \label{fig:ma_env}
\end{figure} 

In this example we recreate the Cooperative Navigation environment from MADDPG \cite{maddpg}, and train to convergence using multi-agent soft actor-critic with the AI Arena framework. In this version of the environment, there are 3 agents that can move in 2D space and must navigate to cover three targets. Targets provide a reward of +1 if they are occupied, so the optimal behavior is to have each entity travel to a unique target, such that all targets are occupied. There is no penalty for colliding with other agents. The environment runs for 300 steps, so the maximum theoretical score is 900 (all entities starting directly on a target and staying there for the duration, for 300x3=900 points). Our implementation of Multi-agent Soft Actor Critic (MASAC) is a direct extension of soft actor critic \cite{sac} to the multi-agent domain using the common critic framework initially described by MADDPG \cite{maddpg}.  MASAC can naturally be distributed. We run with 8 copies of the environment with 8 workers, each worker coordinating three entities in its respective environment and training with a common critic, sharing gradients among all workers at each update.

\begin{figure}[ht!]
\centering
  \includegraphics[width=0.8\linewidth]{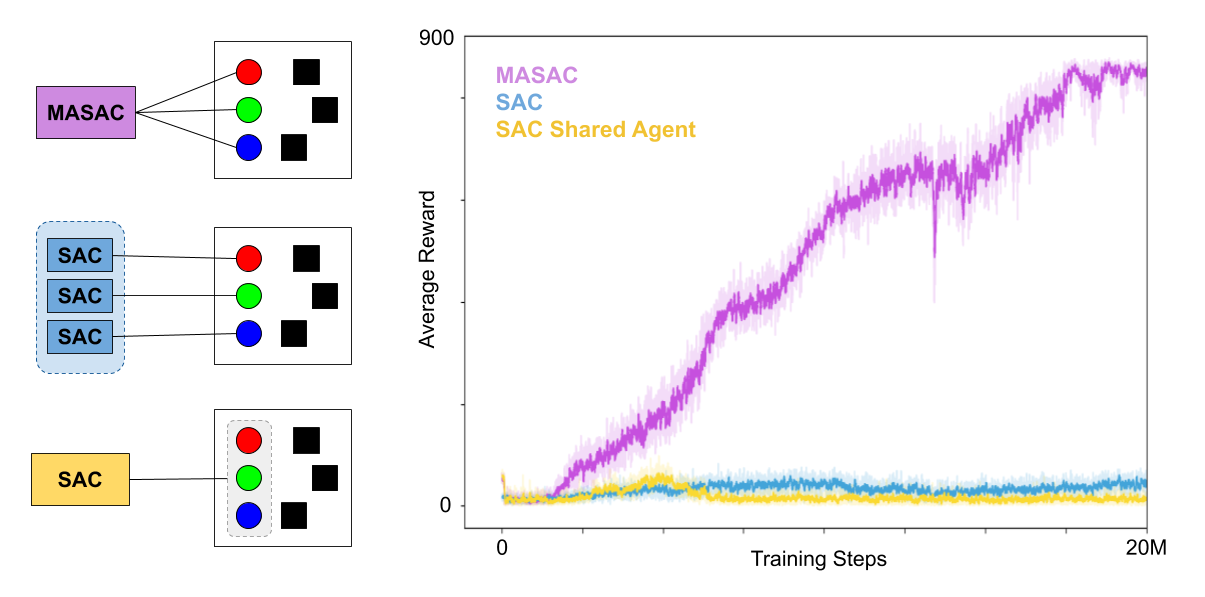}
  \captionsetup{width=0.95\linewidth}
  \caption{\small \emph{Training Curves for MASAC and Comparisons.} \textbf{Left:} Entity assignments for the three approaches: Truly multiagent policy (MASAC), treating each entity as an SAC worker, and grouping all entities into a single agent. In all cases, the assignment was duplicated over several environments for distributed training. \textbf{Right:} The corresponding training curves for each approach. MASAC was the only successful algorithm, making slow and deliberate progress for roughly 18 million steps before levelling off.}
  \label{fig:ma_results}
\end{figure}

As seen in Figure \ref{fig:ma_results}, our agents converge to a cooperative set of behaviors that clear 800 points on average, which is nearly optimal. The agents move quite quickly in this environment, but we have slowed them down in testing to create visualizations of their movements in Figure \ref{fig:ma_env}. While they do not always attempt to reach the nearest target, they have coordinated in such a way that all the targets become occupied. Crucially, they do not communicate during testing, so it is only through training with the common critic that they have learned complementary policies that can deploy independently while still working together.

\begin{figure}[ht!]
\centering
  \includegraphics[width=\linewidth]{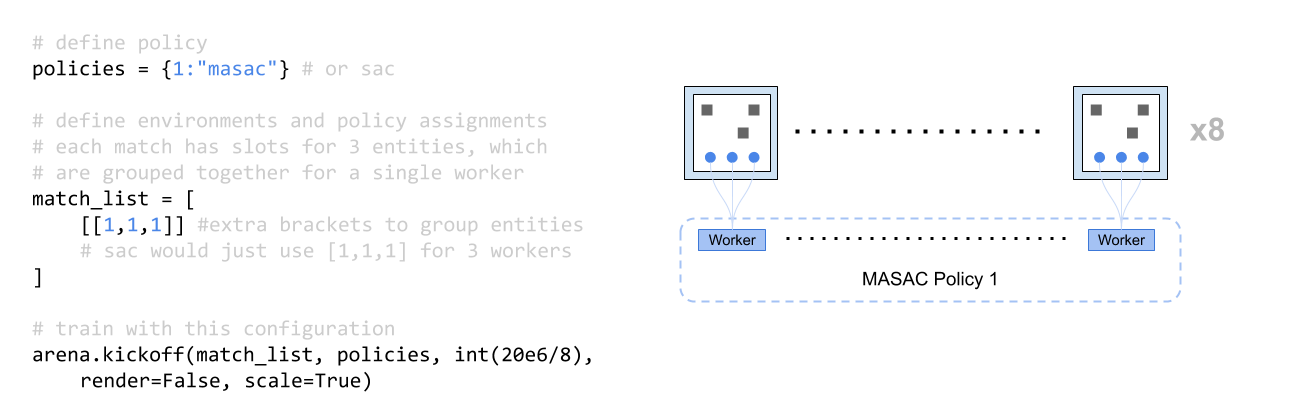}
  \captionsetup{width=0.95\linewidth}
  \caption{\small \emph{Python code and process diagram for a multi-agent algorithm.} \textbf{Left:} Core code snippet needed to setup and run this training scheme. To route multiple entities to a single worker (as needed for a common critic algorithm), the entities are simple grouped by square brackets. Here we indicate that all entities 1-3 correspond to one worker for policy 1. \textbf{Right:} The resulting processes and their organization. In our experiments, eight environments were created, each having three entities but only one worker.}
  \label{fig:ma_code}
\end{figure}

As points of comparison, we note some other approaches to this multi-agent problem. The three entities in the environment could alternatively be treated as separate workers for a single-agent SAC algorithm (comparison 1), or the entities could be grouped together into a single agent that has a single state and accepts one large action (comparison 2, may require slight environment modifications). These comparisons are further described in Figure \ref{fig:ma_results}.

Neither of these comparisons were successful in our experiments. Treating the agents as separate workers for SAC does not properly assign rewards to the agents, since all three agents are collectively rewarded based on target occupancy, and therefore the distributed SAC approach is not able to solve the credit-assignment problem among multiple workers. Treating all three entities as a single agent may suffer from a similar problem in that any rewards that are experienced do not reflect credit for the action taken but rather a subset of the action taken. Also note that it is impossible to have identical data collection among the three approaches. MASAC ran on 8 environments (8 workers, each controlling 3 entities), SAC ran on 6 environments (18 workers, 3 per environment) and the combined entity SAC ran on 8 environments (8 workers, each controlling 3 entities as one agent).

This example demonstrates compatibility with “true” multi-agent algorithms and the ability to scale them. These approaches can be used alongside single-agent algorithms at scale; it is perfectly acceptable to have some entities coordinated in a multi-agent fashion while others are treated as independent workers of a single-agent policy.

\section{CONCLUSIONS}
In this work, we introduce the AI Arena: a scalable framework with flexible abstractions for distributed multi-agent reinforcement learning. Our aim is to help enable RL research for the class of applications that involve multiple teams of agents where each team may have unique learning strategies and where agents within a given team may have localized views of the environment.  Our unique abstractions allow the single agent focused OpenAI Gym interface to be used without modification for multi-agent environments from the perspective of learning algorithms.  We demonstrate the expressiveness of the abstractions for training with multiple environmental configurations simultaneously in the TanksWorld domain.  We compared multiple strategies for multi-agent training in the cooperative navigation domain.  The AI Arena framework software is available as open source software on github at \emph{https://github.com/cgrivera/ai-arena}. The Tanksworld environment for AI safety research  is available at \emph{https://github.com/cgrivera/ai-safety-challenge}.

\bibliographystyle{plain}  
\bibliography{references}  

\appendix

\section{Appendix}
\subsection{Advanced Examples}
The simplicity of the AI Arena interface allows for orchestration of high-level training schemes for which the inner loop is an entire DRL training run. Here we demonstrate the usefulness of this approach towards a population-based training example for Atari Pong.

\subsubsection{Population-based Methods}

\begin{figure}[ht!]
\centering
  \includegraphics[width=\linewidth]{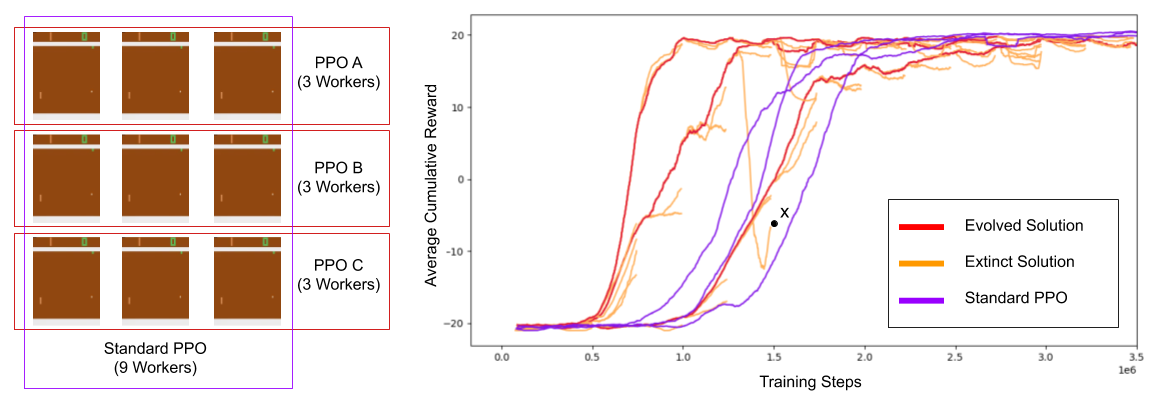}
  \captionsetup{width=0.95\linewidth}
  \caption{\small \emph{Evolution over PPO on Atari Pong.} Nine environments are broken into groups of three to train three independent PPO policies. Periodically, the best policy of the three is copied over to replace the other two, forming a simple round-based evolution scheme. This is compared with using all nine environments to train a single policy instance. \textbf{Left:} Two different assignments of environments to policies, showing the evolution case (red) and the typical use case (purple) \textbf{Right:} Three training runs of each approach, showing evolution in red and typical PPO in purple. Orange "hairs" visualize policies that were not selected to continue at the end of a generation (every 250k training steps). Point X shows an especially poor performing policy that was eliminated via evolution.}
  \label{fig:evo}
\end{figure}

The AI Arena paradigm of allowing arbitrary connections between entities and policies can provide for some interesting and potentially powerful training schemes. Here we demonstrate a toy variation on PPO in which three policies are maintained instead of one, and this population of three is evolved over many rounds of training on the Atari game Pong. We use nine environment instances such that each PPO instance is a small distributed learner using three of the nine environments. Every 250,000 training steps, the policies are paused and only the most performant is allowed to continue by copying its network weights over to the two lesser policies. For comparison, we also examine the standard approach of applying all computational resources to a single policy (all nine environments to one policy). Results from both approaches are shown in Figure \ref{fig:evo}.

While a simple example, this demonstrates the ability to set up complex training schemes that may involve populations of agents, multiple rounds of training, and many environments. A population-based approach may not result in performance gains on Pong, but it provides another option for training which may be difficult to set up in other software frameworks. In particular, this training scheme shows some interesting properties that can be seen in Figure \ref{fig:evo}. Namely, two of the three runs converge with less data than their standard counterparts (although not necessarily less network updates, as they have less data per update). The population-based approach also seems to protect against irregularities in training, such as poor initialization or catastrophic updates, by weeding out poor performing policies. One population member had an especially poor training generation in one of the runs (point X in Figure \ref{fig:evo}), which was abandoned in the next generation. If this had been a single PPO run, this catastrophic performance drop could have long-term effects. In a bit of irony, the purpose of the PPO algorithm is to clip network updates such that these performance hits are avoided in the first place, but it is not infallible.

\end{document}